\documentclass[runningheads]{llncs}
\usepackage[T1]{fontenc}
\usepackage{lmodern}
\usepackage{graphicx}
\usepackage{amssymb}
\usepackage{amsmath}
\usepackage{subcaption}
\usepackage{algorithm}
\usepackage{algorithmic}
\begin{document}
\raggedbottom
\title{Active learning of digenic functions with boolean matrix logic programming}
%
%
\author{Lun Ai\inst{1} \and Stephen H. Muggleton\inst{2} \and
Shi-Shun Liang\inst{1} \and
Geoff S. Baldwin\inst{1}}
\authorrunning{L. Ai et al.}
%
\institute{Department of Life Sciences \and
Department of Computing \\
Imperial College London, London, UK\\
\email{\{lun.ai15,s.muggleton,shishun.liang20,g.baldwin\}@imperial.ac.uk}}
\maketitle              
\begin{abstract}
We apply logic-based machine learning techniques to facilitate cellular engineering and drive biological discovery, using a comprehensive knowledge base of metabolic processes called a genome-scale metabolic network model (GEM). Predicted host behaviours are not always correctly described by GEMs. Learning the intricate genetic interactions within GEMs presents computational and empirical challenges. To address these difficulties, we describe a novel approach called Boolean Matrix Logic Programming (BMLP) by leveraging boolean matrices to evaluate large logic programs. We introduce a new system, $BMLP_{active}$, which efficiently explores the genomic hypothesis space by guiding informative experimentation through active learning. In contrast to sub-symbolic methods, $BMLP_{active}$ encodes a state-of-the-art GEM of a widely accepted bacterial host in an interpretable and logical representation using datalog logic programs. Notably, $BMLP_{active}$ can successfully learn the interaction between a gene pair with 90\% fewer training examples than random experimentation, overcoming the increase in experimental design space. $BMLP_{active}$ enables rapid optimisation of metabolic models and offers a realistic approach to a self-driving lab for microbial engineering.

\end{abstract}


\section{Introduction}

Our study explores the applicability of abductive reasoning and active learning to extensive biological systems. We expand the scope of logic-based gene function learning by looking at a genome-scale metabolic model. The genome-scale metabolic model (GEM), iML1515 \cite{iML1515}, encompasses 1515 genes and 2719 metabolic reactions of the \textit{Escherichia coli} (\textit{E. coli}) strain K-12 MG1655, a versatile host organism in metabolic engineering to produce valuable compounds. In iML1515, interactions between gene pairs have been identified as a key error source \cite{bernstein_evaluating_2023}. Learning these interactions requires exploring a combinatorial hypothesis space. Given that biological relationships are commonly described logically, Inductive Logic Programming (ILP) \cite{ILP1991} is particularly adept at operating on biological knowledge bases. We propose Boolean Matrix Logic Programming (BMLP) where we employ boolean matrices as underlying bottom-up evaluation mechanisms of large datalog programs. This approach allows us to classify phenotypic effects using a datalog encoding of the state-of-the-art GEM iML1515. We focused on re-discovering the function associated with the key gene $tyrB$ in the Tryptophan biosynthesis pathways. Overlapping functions between two genes $tyrB$ and $aspC$ is a digenic interaction since they are responsible for producing the same amino acid. To learn from a combinatorial hypothesis space with high data efficiency, we create an active learning system $BMLP_{active}$ to learn this digenic interaction, requiring 90\% fewer experiments than random experimentation. Abductive reasoning and active learning via ILP were successfully demonstrated in biological discovery by the prominent Robot Scientist \cite{King04:RobotScientist}. However, this demonstration was limited to only 17 genes in the aromatic amino acid pathways of yeast. In contrast, $BMLP_{active}$ is the first logic programming system to learn digenic interaction on the genome scale from GEMs.

\begin{figure}[t]
      \centering
      \includegraphics[width=\textwidth]{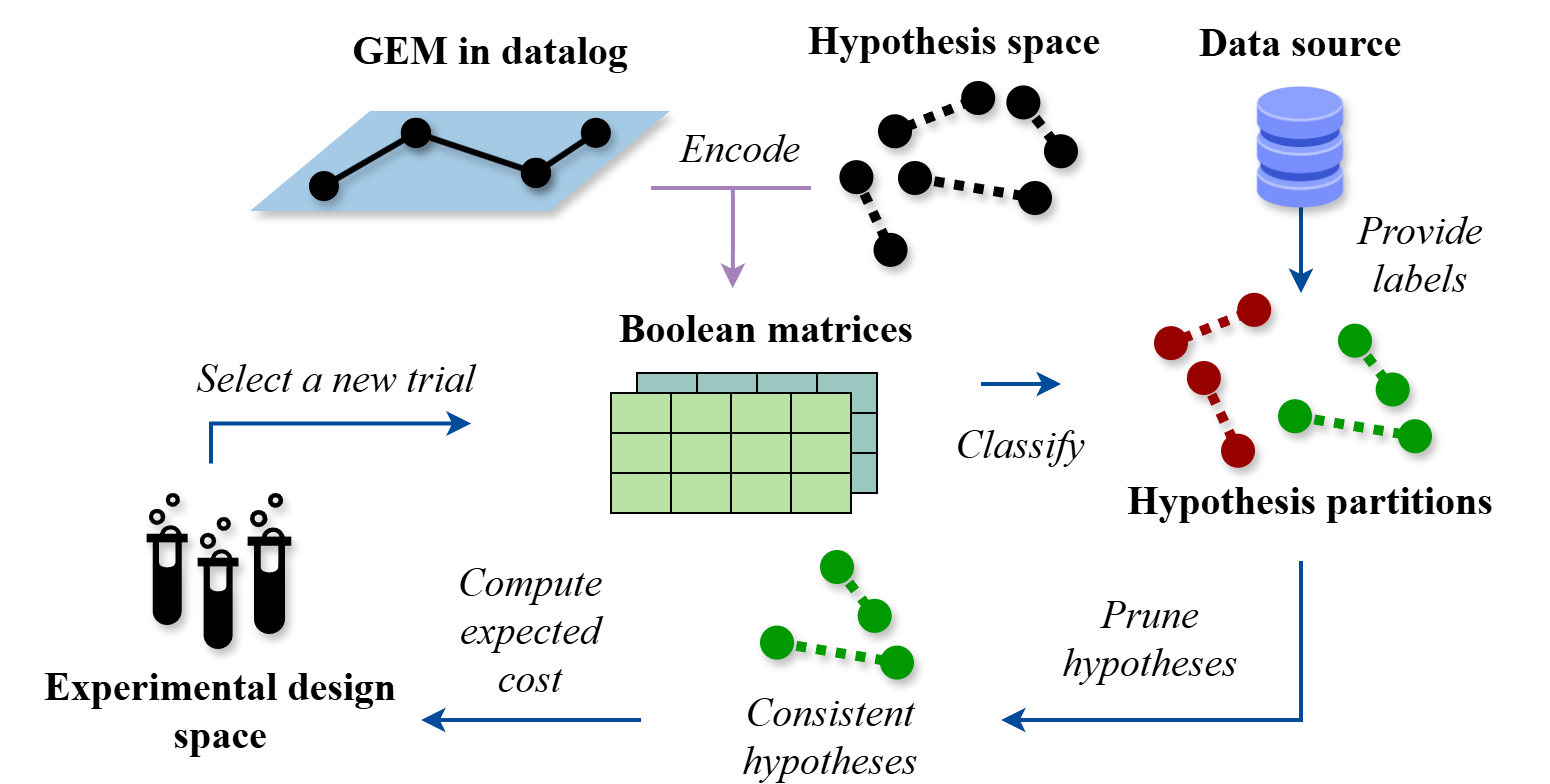}
      \caption{Certain genetic mutations would block pathways, causing cells to die (positive label). $BMLP_{active}$ finds a gene-reaction association hypothesis to explain the pathway blockage and lethality. It encodes the GEM iML1515 as boolean matrices and uses them to classify genetic mutation experiment labels for every hypothesis. It consults a data source to request ground truth labels. $BMLP_{active}$ iteratively refutes hypotheses inconsistent with the labels. }
      \label{fig:framework}
    \end{figure}


\section{Related work}

Many computational scientific discovery systems \cite{langley_scientific_1987,todorovski_integrating_2006,brunton_discovering_2016,guimera_bayesian_2020,petersen_deep_2020} were designed to formulate symbolic hypotheses from experimental results. One cannot directly employ these systems for experimental planning. In contrast, the Robot Scientist \cite{King04:RobotScientist,king_automation_2009} automatically proposed hypotheses, designed experiments and performed them using laboratory robotics. Experiments were actively selected to minimise the expected cost of experimentation for learning gene functions in yeast \cite{ase_progol}. Notably, we apply $BMLP_{active}$ to the GEM iML1515 which represents a significant increase in complexity compared to the aromatic amino acid pathways investigated by the Robot Scientist platforms.

BMLP uses boolean matrices to compute the least model. Obtaining the least Herbrand model of linear recursive datalog programs can be reduced to computing the transitive closure of boolean matrices \cite{peirce_collected_1932,copilowish_matrix_1948}. Fischer and Meyer \cite{fischer_boolean_1971} and Ioannidis \cite{ioannidis_computation_1986} showed a divide-and-conquer boolean matrix computation technique by viewing relational databases as graphs. An ILP system called DeepLog \cite{muggleton_hypothesizing_2023} employed boolean matrices for choosing optimal background knowledge to derive a compact bottom clause. In contrast to BMLP, a great body of work \cite{lin_satisability_2013,grefenstette_towards_2013,sato_linear_2017,cohen_tensorlog_2020,sato_boolean_2021,sato_differentiable_2023} only studied approximated evaluations of logic program in tensor spaces. Solving certain recursive programs is difficult whereas these can be solved by an iterative bottom-up evaluation approach \cite{sato_linear_2017} such as BMLP. 

\section{Modelling a GEM with a Petri net}

In an auxotrophic mutant experiment, genes are deleted from a cell. Thus, key metabolic reactions might be no longer viable due to the gene removal, leading to insufficient cell growth. We model the metabolic mechanisms using a GEM. A GEM contains biochemical reactions of substances. Each reaction involves chemical substances $x_i$ and $y_j$: 
\begin{flalign}
    \text{Irreversible: } x_1 + x_2 + ... + x_m &\longrightarrow y_1 + y_2 + ... + y_n \nonumber\\
    \text{Reversible: } x_1 + x_2 + ... + x_m &\longleftrightarrow y_1 + y_2 + ... + y_n \nonumber
\end{flalign}
which is often represented by a Petri net \cite{sahu_advances_2021}.  The reactions in a GEM are transitions (edges in a Petri net) between reactants and products (nodes in a Petri net). We associate \textit{binary} phenotypic effects -- normal or reduced cell growth of auxotrophic mutants compared to the wild-type, with reachability to key substances in the metabolic network. To evaluate this reachability problem, we look at Petri nets with restrictions, namely \textit{one-bounded elementary nets} (OEN)  \cite{rozenberg_elementary_1998} where nodes are marked with at most one token (Example \ref{ex:transformation}). The reachability problem is evaluated as a recursive datalog program with two arguments. Each argument of our transformed datalog program concatenates multiple constants, e.g. $h2\_o2$. This results in a hypergraph \cite{berge_hypergraphs_1989}, a directed graph where an edge can connect to any number of nodes. To construct this datalog program, we consider a hypergraph that describes the association between substances and reactions.
\begin{example}
\textit{$\mathcal{P}_1$ represents a metabolic network as an OEN and hypergraph. The clauses $reaction\_1$, $reaction\_2$ and $reaction\_3$ are the three transitions $t_1$, $t_2$ and $t_3$. $c1\_c2$ and $c3\_c4$ are two hypernodes. Both graphs show that the node $c_5$ is reachable from $\{ c_1, c_2 \}$ via $t_1$ and $t_2$. We can evaluate the groundings of $pathway(m1, Y)$ in the Herbrand model which includes $pathway(m1, c5)$.  The OEN and hypergraph are illustrated below: }\\
\begin{minipage}{0.6\textwidth}
\begin{gather}
\mathcal{P}_1:\left\{
\begin{aligned} 
    metabolites(m1,c1\_c2). &\\
    metabolites(m1,c1).&\\
    metabolites(m1,c2).&\\
    reaction\_1(c1\_c2, c3\_c4). &\\
    reaction\_1(c1\_c2, c3). &\\
    reaction\_1(c1\_c2, c4).&\\
    reaction\_2(c3\_c4, c5). &\\
    reaction\_3(c5, c4). &\\
    reaction(X,Y) \gets metabolites(X,Y). &\\
    reaction(X,Y) \gets reaction\_1(X,Y). &\\
    reaction(X,Y) \gets reaction\_2(X,Y). &\\
    reaction(X,Y) \gets reaction\_3(X,Y). &\\
    pathway(X,Y) \gets reaction(X,Y). &\\ 
    pathway(X,Y) \gets reaction(X,Z), &\\
                pathway(Z,Y). & \nonumber
\end{aligned}
\right\}
\end{gather} 
\end{minipage}
\begin{minipage}{0.4\textwidth}
    \centering
    \includegraphics[width=0.9\linewidth]{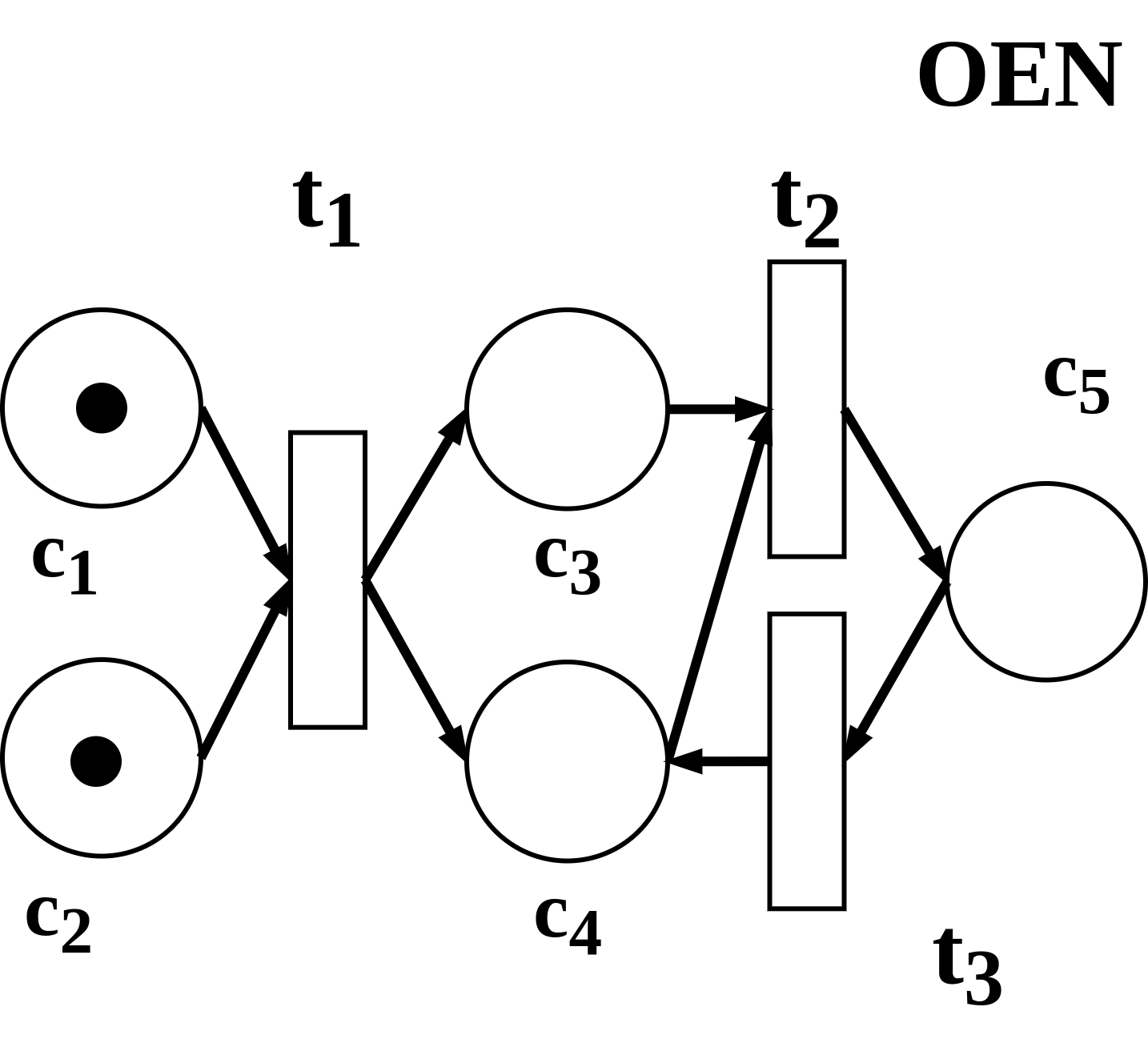}
    \includegraphics[width=0.9\linewidth]{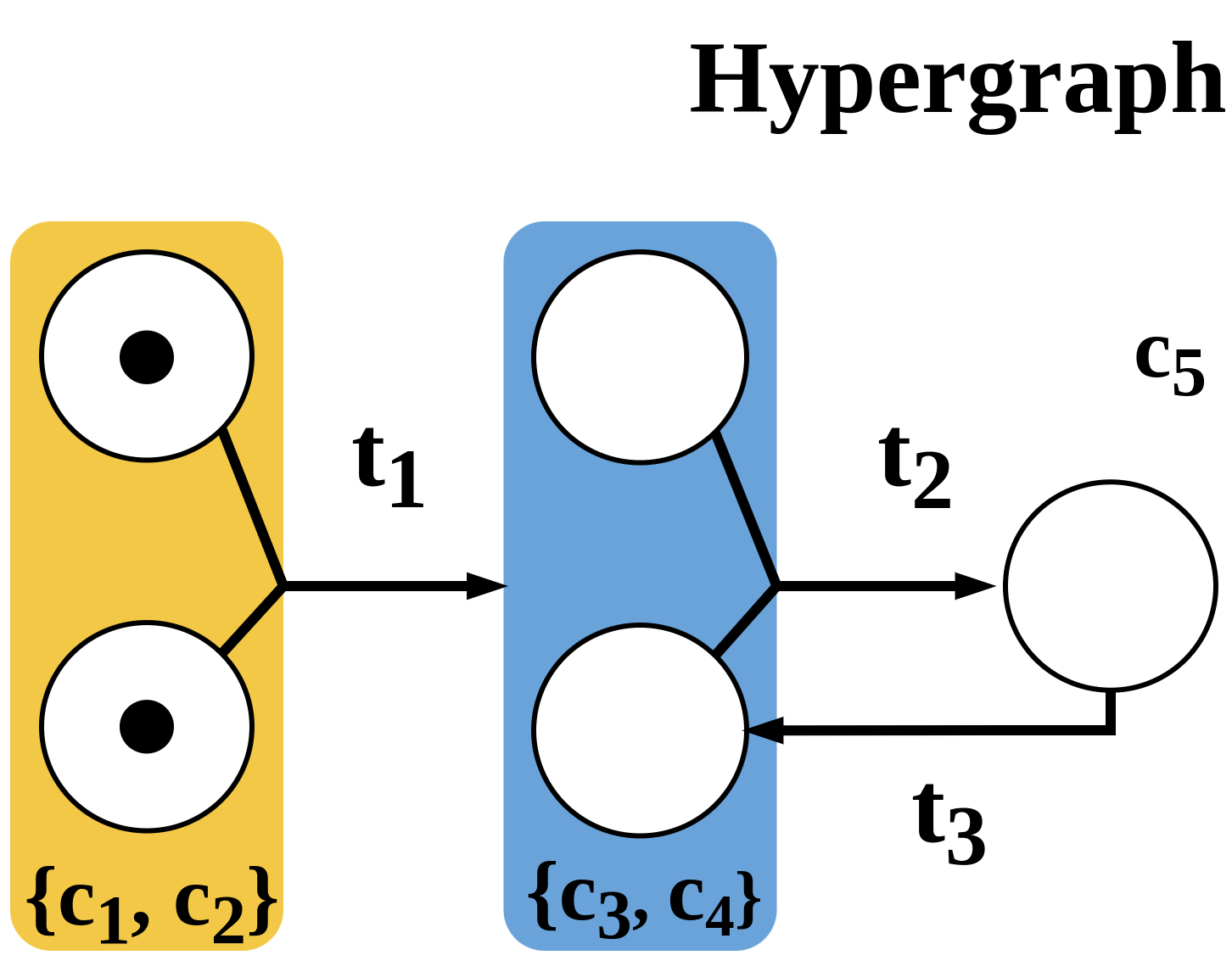}
    \label{fig:enter-label}
\end{minipage}
\label{ex:transformation}
\end{example}

\section{Boolean matrix logic programming}
\label{sec:framework_bmlp}
We propose the Boolean Matrix Logic Programming (BMLP) problem. In contrast to traditional logic program evaluation, BMLP uses boolean matrices to evaluate recursive datalog programs with arity at most two (at most two arguments) and at most two body literals, namely the $H_2^2$ program class \cite{muggleton_meta_interpretive_2015}. The $H_2^2$ program class has the Universal Turing Machine expressivity when extended with functional symbols  \cite{tarnlund_horn_1977}. 

\begin{definition} [Boolean Matrix Logic Programming (BMLP) problem]
    Let $\mathcal{P}$ be a datalog program containing a set of clauses with the predicate symbol $r$. The goal of Boolean Matrix Logic Programming (BMLP) is to find a boolean matrix $\textbf{R}$ encoded by a datalog program such that $(\textbf{R})_{i,j}$ = 1 if $\mathcal{P} \models r(c_i, c_j)$ for constants $c_i, c_j$ and $(\textbf{R})_{i,j}$ = 0 otherwise.
\end{definition}

\begin{figure}[t]
    \centering
    \begin{tabular}{cc}
        \begin{subfigure} {0.6\textwidth}
            \includegraphics[width=\linewidth]{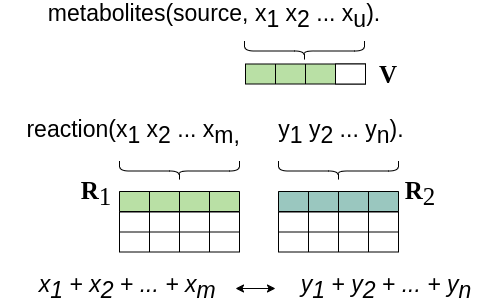}
            \caption{}
        \end{subfigure} & 
        \begin{subfigure} {0.3\textwidth}
            \includegraphics[width=\linewidth]{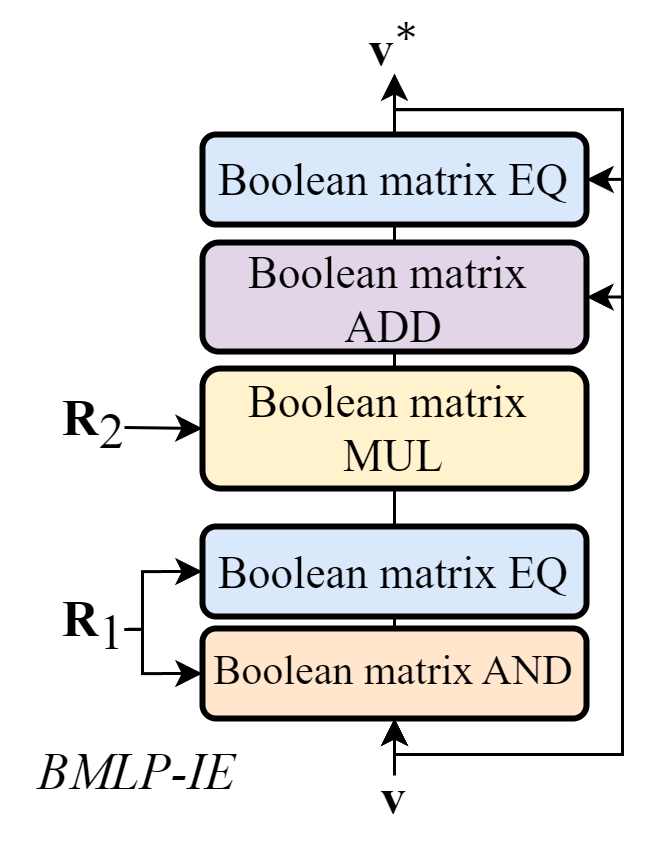} 
            \caption{}
        \end{subfigure} 
    \end{tabular}
    \caption{(a) The vector $\textbf{v}$ encodes source chemical metabolites. All reactions are represented in the boolean matrices $\textbf{R}_1$ and $\textbf{R}_2$. (b) The module BMLP-IE computes $\textbf{v}^*$, the closure of reaction products, using binary AND over rows and boolean matrix addition (ADD), multiplication (MUL) and equality (EQ) \cite{copilowish_matrix_1948}. }
    \label{fig:bmlp_ie}
\end{figure}

Consider a ground term $u$ that concatenates $n$ constant symbols $c_1, c_2, ..., c_n$. We can represent $u$ using a $n$-bit binary vector. When $c_k$ is in $u$, the $k$-th bit of the binary vector is 1. All reactions in iML1515 are encoded this way to form rows in two boolean matrices $\textbf{R}_1$ and $\textbf{R}_2$ (Fig \ref{fig:bmlp_ie}a).

We implement a BMLP module in SWI-Prolog \cite{wielemaker:2011:tplp} called iterative extension (BMLP-IE) for the task of bottom-up evaluation of reachability in metabolic networks. Each binary vector is represented as an integer. BMLP-IE (Fig \ref{fig:bmlp_ie}b) iteratively expands the set of facts derivable from a partially grounded query $r(u, Y)$. For $\mathcal{P}$ containing $n$ constant symbols, we represent $u$ as a $1 \times n$ row vector $\mathbf{v}$. $\textbf{R}_1$ and $\textbf{R}_2$ can be multiplied with boolean vectors to turn transitions ``on'' or ``off' in the OEN to modify the network connectivity. This module gives a 170x runtime boost for this task compared to SWI-Prolog \cite{ai_boolean_2024}, showing the benefit of a special-purpose module.


\section{Active learning}
\label{sec:framework_active_learning}

Our active learning system $BMLP_{active}$ selects experiments to minimise the expected value of a user-defined cost function and iteratively prunes hypotheses inconsistent with experimental outcomes. $BMLP_{active}$ uses the compression score of a hypothesis $h$ \cite{ase_progol} based on the Minimal Description Length principle \cite{conklin_complexity-based_1994} to compute the posterior probability:
\begin{flalign*}
    compression(h,E) &= |E^+| - \frac{|E^+|}{pc_h} (size(h) + fp_h)\\
    p'(h|E) &= \frac{2^{compression(h, E)}}{\sum_{h_i\in H} 2^{compression(h_i, E)}}
\end{flalign*}

$E$ are labelled examples and $E^+$ are positive examples. $pc_h$ and $fp_h$ are the number of positive predictions and false positive coverage. This compression function favours a general hypothesis with high coverage and penalises over-general hypotheses that incorrectly predict negative examples. $BMLP_{active}$ uses the following heuristic function \cite{ase_progol} to select an experiment from candidate experiments $T$ with an approximated minimum expected cost:
\begin{flalign*}
    EC(H,T) \approx min_{t \in T} \, [C_t &+ p(t) (mean_{t' \in T - \{t\}} C_{t'}) J_{H_{t}} \\ \nonumber
    & + (1 - p(t)) (mean_{t' \in T - \{t\}} C_{t'}) J_{\overline{H_{t}}}]
\end{flalign*}
$H_{t}$ and $\overline{H_{t}}$ are subsets of hypotheses $H$ consistent and inconsistent with $t$'s label. $J_{H_{t}} and J_{\overline{H_{t}}}$ are calculated according to the entropy $ - \sum_{h \in H} p'(h|E) \, log_2 (p'(h|E))$ where $H$ is replaced by $H_{t}$ or $\overline{H_{t}}$. The probability $p'(h|E)$ is calculated from the compression function. $p(t)$ is the probability that the label of the experiment $t$ is positive and is computed by the probability sum of consistent hypotheses $\sum_{h \in H_{t}} p'(h|E)$. $C_t$ is the cost of $t$ from a user-defined experiment cost function. 

For some hypothesis space $H$ and background knowledge $BK$, $V_s$ is the version space consistent with $s$ training examples with the assumption that an active version space learner selects one instance per iteration. The shrinkage of the hypothesis space is $\frac{|V_{s+1}|}{|V_s|}$ after querying the label of the $s$-th instance. The minimal reduction ratio $p(x_s, V_s)$ is the minority ratio of the version space $V_s$ partitioned by an instance $x_s$.
\begin{flalign*} 
    p(x_s, V_s) = \frac{min(\, |\{h \in V_s | \,h \cup BK \models x_s\}|, |\{h \in V_s | \,h \cup BK \not\models x_s\}|)}{|V_s|} 
\end{flalign*}
The optional selection strategy is to select instances with minimal reduction ratios as close as possible to $\frac{1}{2}$ to eliminate up to $50\%$ of the hypothesis space \cite{mitchell_generalization_1982}. A passive learner is considered a learner using random example selection since it does not have control over which training examples it uses. 

\begin{theorem} (Active learning sample complexity bound \cite{ai_boolean_2024})
    For some $\phi \in [0, \frac{1}{2}]$ and a small hypothesis error $\epsilon > 0$, if an active version space learner can select instances to label from an instance space $\mathcal{X}$ with minimal reduction ratios greater than or equal to $\phi$, the sample complexity $s_{active}$ of the active learner is 
    \begin{flalign}
        s_{active} \leq \frac{\epsilon}{\epsilon + \phi} s_{passive} + c
        \label{eq:sample_complexity_bound}
    \end{flalign}
    where c is a constant and $s_{passive}$ is the sample complexity of learning from randomly labelled instances.
    \label{theorem:active_learning}
\end{theorem}

Theorem \ref{theorem:active_learning} says that the number of instances needed by active learning should be always smaller than the number of randomly sampled examples given a positive $\phi$. Recent work \cite{ai_boolean_2024} has theoretically and empirically supported Theorem \ref{theorem:active_learning}. This means the biological experiment sequence needed to arrive at a finding can be shortened by active learning. 

We applied BMLP-IE to obtain predicted phenotypic effects for all combinations of candidate experiments and hypotheses. These predictions were compared with experimental data. Labels of experiment instances were requested from a data source, e.g. a laboratory or an online dataset. We have only considered labels from synthetic or online phenotype data. However, $BMLP_{active}$ can request data from a laboratory to automate experiments. 


\section{Experiments}
\label{sec:experiments}
\begin{figure}[t]
\centering
\includegraphics[width=0.9\textwidth]{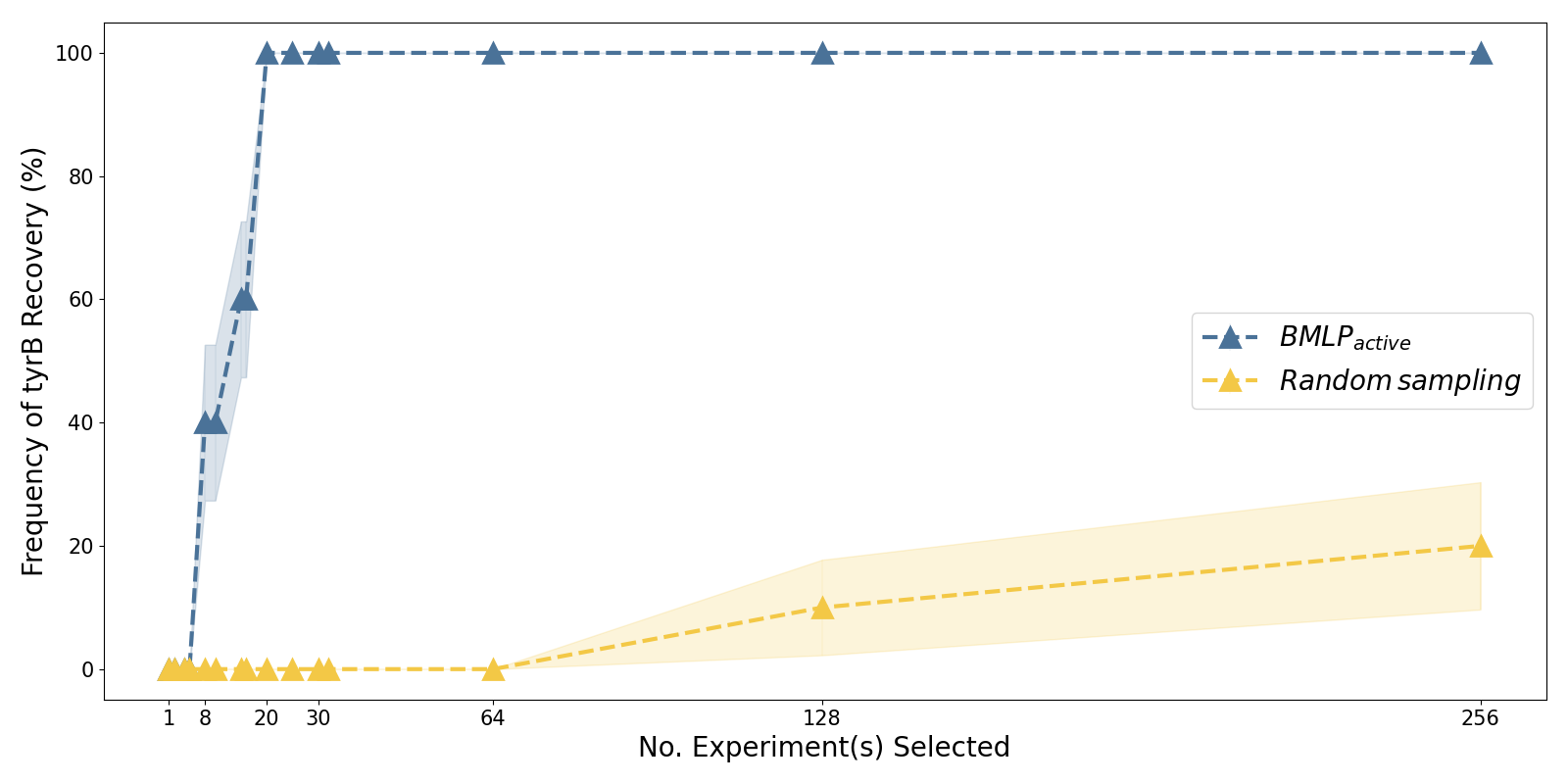}
\caption{tyrB isoenzyme function recovery frequency. The experimental space had $(\binom{33}{2} + 33) \times 7 = 3927$ instances for double gene-knockout synthetic data and single gene-knockout experimental data of key 33 genes in 7 conditions. The hypothesis space contained $(27 \times 32 + 6 \times 31) + 2 = 1052$ candidate gene-enzyme associations related to 27 single-function genes, 6 double-function genes, the tyrB original function and an empty hypothesis.}
\label{fig:isoenzyme_recovery}
\end{figure} 

We removed the metabolic reaction associated with $tyrB$ from iML1515. This GEM model iML1515 \cite{iML1515} was first represented as a datalog program and then as boolean matrices for computation. The random selection strategy randomly sampled $N$ instances from the instance space. $BMLP_{active}$ selected $N$ experiments from this instance space to actively learn from the hypothesis space. Both methods output the hypothesis with the highest compression. 

In Fig \ref{fig:isoenzyme_recovery} we observed that $BMLP_{active}$ successfully guarantees the recovery of the correct gene function with as few as 20 experiments. Random sampling failed to effectively prune competitive hypotheses even with 250+ experiments. This result also shows that $BMLP_{active}$ significantly reduced the number of experiments needed by 90\% compared to random experiment sampling. 

\section{Conclusion and future work}

We explored abductive reasoning and active learning using Boolean Matrix Logic Programming (BMLP), which utilises boolean matrices encoded in SWI-Prolog to compute datalog programs. We applied our active learning system $BMLP_{active}$ to learn a key digenic function in a state-of-the-art genome-scale metabolic network. Though we focused on a reduced set of hypotheses in the experiment, the combinatorial space was sufficiently large that random experiment selection was not viable as an experimentation strategy in a discovery process. The remarkable increase in the data efficiency of $BMLP_{active}$ demonstrates its potential even as the experimental design space grows exponentially. 

Petri nets complement knowledge representation with dynamic analysis \cite{sahu_advances_2021}. However, the simulation of Petri nets in logic programming has been much less explored. Future work will explore the link between Petri nets and Probabilistic Logic Programming \cite{de_raedt_probabilistic_2015}. Transitions could have firing likelihood constraints and this uncertainty can be modelled by probabilistic logic programs. \\

\begin{credits}
\subsubsection{\ackname} The first, third and fourth authors acknowledge support from the UKRI 21EBTA: EB-AI Consortium for Bioengineered Cells and Systems (AI-4-EB) award (BB/W013770/1). The second author acknowledges support from the UK’s EPSRC Human-Like Computing Network (EP/R022291/1), for which he acts as Principal Investigator. 
\end{credits}
%
%
%
\bibliographystyle{splncs04}
\bibliography{active_abduction}
\end{document}